\documentclass[letterpaper, 10 pt, conference]{ieeeconf}
\IEEEoverridecommandlockouts      

\usepackage{times}

\makeatletter
\let\NAT@parse\undefined
\makeatother
\usepackage[numbers,compress]{natbib}

\usepackage{multicol}
\usepackage[bookmarks=true]{hyperref}
\usepackage{amsmath} 
\usepackage{amssymb} 
\usepackage{algorithm}
\usepackage{algpseudocode}
\usepackage{graphicx}
\usepackage{array}
\usepackage{booktabs}
\usepackage{caption}
\usepackage{subcaption}
\usepackage{xcolor}
\usepackage{multirow}
\usepackage[skip=2pt,font=small]{caption}

\usepackage{makecell} 
\usepackage{paralist}
\usepackage{mathrsfs}

\definecolor{Red}{rgb}{1,0,0}

\newcolumntype{P}[1]{>{\centering\arraybackslash}p{#1}}
\newcolumntype{M}[1]{>{\centering\arraybackslash}m{#1}}

\definecolor{cameraReadyColor}{rgb}{0,0,0}
\makeatletter
\newcommand*{\cameraReady}{\@ifnextchar\bgroup{\cameraReady@}{\color{cameraReadyColor}}}
\newcommand*{\cameraReady@}[1]{{\textcolor{cameraReadyColor}{#1}}}

\begin{document}

\title{\LARGE \bf Diff-DAgger: Uncertainty Estimation with \\Diffusion Policy for Robotic Manipulation}

\author{Sung-Wook Lee$^{1}$, Xuhui Kang$^{1}$, Yen-Ling Kuo$^{1}$
  \thanks{$^{1}$University of Virginia\newline {\tt\small $\phantom{0}$ \{dcs3zc,xuhui,ylkuo\}@virginia.edu}}
}



%

\maketitle

\begin{abstract}
Recently, diffusion policy has shown impressive results in handling multi-modal tasks in robotic manipulation. 
However, it has fundamental limitations in out-of-distribution failures that persist due to compounding errors and its limited capability to extrapolate. 
One way to address these limitations is robot-gated DAgger, an interactive imitation learning with a robot query system to actively seek expert help during policy rollout. 
While robot-gated DAgger has high potential for learning at scale, existing methods like Ensemble-DAgger struggle with highly expressive policies: They often misinterpret policy disagreements as uncertainty at multi-modal decision points.
To address this problem, we introduce Diff-DAgger, an efficient robot-gated DAgger algorithm that leverages the training objective of diffusion policy. We evaluate Diff-DAgger across different robot tasks including stacking, pushing, and plugging, and show that Diff-DAgger improves the task failure prediction by 39.0\%, the task completion rate by 20.6\%, and reduces the wall-clock time by a factor of 7.8. We hope that this work opens up a path for efficiently incorporating expressive yet data-hungry policies into interactive robot learning settings. The project website is available at: \href{https://diffdagger.github.io}{https://diffdagger.github.io}.

\end{abstract}

\IEEEpeerreviewmaketitle

\section{Introduction}
Imitation learning for manipulation tasks poses significant challenges due to the need for long-horizon decision making, high precision, and adaptability in dynamic environments. 
Recently, significant advances have been made using offline approaches, also known as behavior cloning, by adopting generative models for action sequence prediction. These models mitigate the compounding error issue by reducing the number of action inferences \cite{zhao2023learning} and addressing multi-modality through generative capabilities \cite{ho2020denoising}. In particular, diffusion policy \cite{chi2023diffusion} trains the model to predict the gradient of the action score function, 
and it has achieved state-of-the-art results for learning different robotic tasks. 

While diffusion policy effectively handles highly multi-modal demonstrations, it still faces a core issue of behavior cloning: the out-of-distribution (OOD) failure, where the policy compounds errors in action prediction and cannot extrapolate in unseen states. A common strategy to address OOD failure is DAgger \cite{ross2011reduction}, an interactive learning method which iteratively refines the policy by incorporating expert interventions. In particular, robot-gated DAgger lets robots decide when to seek expert help, removing the need for constant human supervision and enabling large-scale learning across multiple robots \cite {hoque2023fleet}. The effectiveness of this approach, therefore, relies heavily on the robot's ability to accurately estimate its own uncertainty. 

\begin{figure}[h!]
\centering
    \includegraphics[width=0.95\linewidth]
    {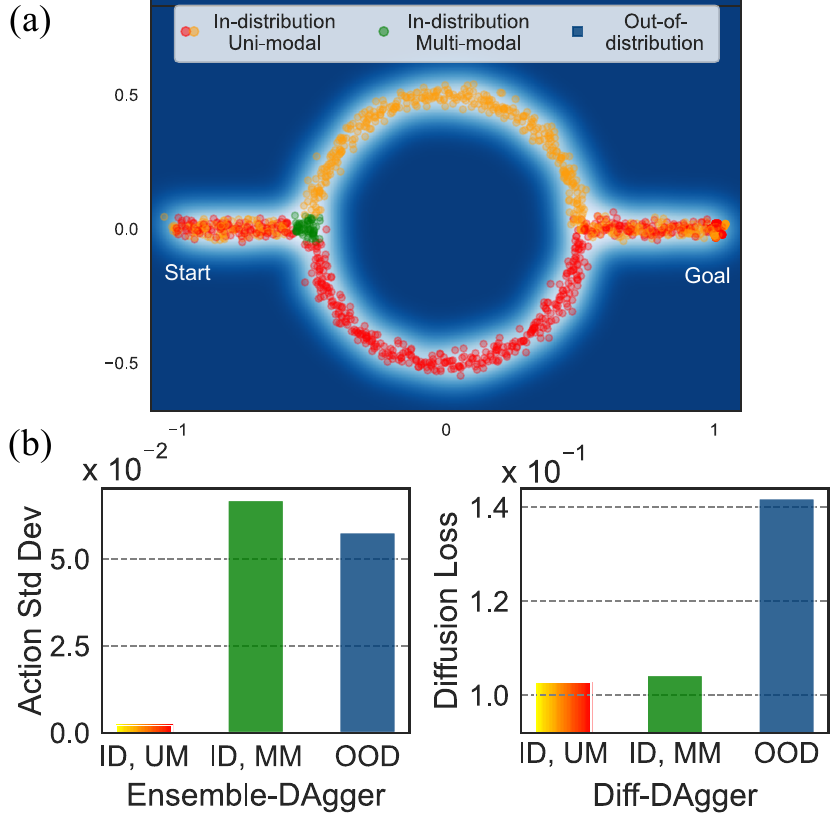}
    \caption{\textbf{(a)} Visualization of 2D navigation toy example. The orange dots represent the mode for following a clockwise path around the circle and the red dots represent another mode for following a counter-clockwise path. The green dots represent the multi-modal region where multiple diverging paths exist. The blue region indicates out-of-distribution.
    \textbf{(b)} Comparison of action standard deviation and diffusion loss for different regions across 100 runs in the 2D navigation example. The left bar plot displays the standard deviation of action across policies for in-distribution uni-modal (ID, UM), in-distribution multi-modal (ID, MM), and out-of-distribution (OOD) regions.
    The right bar plot shows the corresponding diffusion loss for these regions.
}
    \label{fig:toy_example}
\end{figure}

However, none of the existing robot-gated DAgger approaches fully addresses the multi-modality present in demonstration data. For example, the most widely adopted Ensemble-DAgger approaches use the action variance among an ensemble of policies to estimate uncertainty \cite{menda2019ensembledagger, hoque2021thriftydagger, dass2022pato}. This means that when tasks admit multiple solutions, policies might erroneously indicate high uncertainty even when in a previously visited state. 
We illustrate this problem with a 2D navigation example (Figure \ref{fig:toy_example} (a)), where two primary paths—clockwise and counter-clockwise around a circle—lead to the goal. This scenario allows us to categorize regions as in-distribution uni-modal (along the circular paths, shown as mixed red and orange), in-distribution multi-modal (at the path divergence point, shown as green), and out-of-distribution (significant deviations from expected paths, shown as blue). While ensemble methods for robot-gated DAgger effectively distinguish in-distribution uni-modal from out-of-distribution data using action variance, they fail at in-distribution multi-modal points due to policy disagreement (Figure \ref{fig:toy_example} (b)).

\begin{figure*}[t]

    \begin{center}
        {\includegraphics[width=0.98\linewidth]{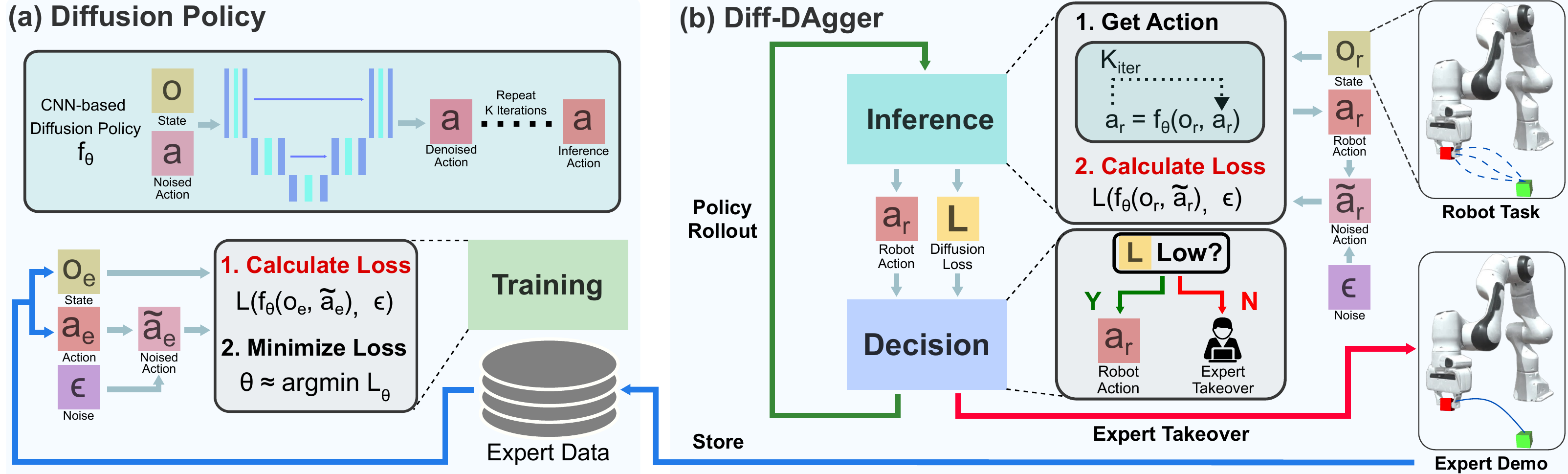} 
        }
        \caption{\textbf{Overview of Diff-DAgger.} \textbf{(a) Diffusion Policy and Training}: In our study, we use the diffusion policy with a U-net architecture, a CNN-based model. During the training phase, diffusion policy is trained on static expert data. \textbf{(b) Diff-DAgger during Deployment}: At each timestep during the online learning phase, the robot performs two steps—\textbf{inference} and \textbf{decision}. In the inference step, the policy makes action inference by iteratively denoising a random noise, conditioned on the current observation. After action generation, the training loss function is used to compute the loss associated with the generated action. In the decision step, the robot proceeds with the action if the loss is low, otherwise it asks the expert to intervene. The arrows indicate the flow of steps.
        }%
    \end{center}
    \label{fig:main}%
\end{figure*}

Incorporating an expressive policy class that handles multi-modal data into the DAgger framework is a promising but under-explored direction. In light of this, we propose Diff-DAgger, an efficient robot-gated DAgger algorithm that addresses multi-modality in demonstration data while reducing wall-clock time compared to other methods. Diff-DAgger uses a query system based on the loss function of the diffusion model, referred to as diffusion loss, which significantly improves the effectiveness of robot queries across various evaluation metrics.
In five challenging manipulation tasks, including two real-world tasks, Diff-DAgger achieves a 39.0\% higher F1 score in predicting task failure, and the collected data leads to 20.6\% improvements in policy performance while improving the wall clock time by a factor of 7.8 in comparison to the baseline methods.

We make the following contributions:
First, we develop a novel robot-gated query system based on diffusion loss to better handle multi-modality in expert demonstrations. 
Second, we quantitatively evaluate Diff-DAgger's efficiency in terms of task failure prediction, task completion, and wall-clock time during interactive learning to demonstrate its superiority over other baseline methods. 
Lastly, we discuss design choices that accelerate policy learning to enable faster convergence and improved overall performance.

\section{Related Work}
\subsection{Interactive Imitation Learning}

The goal of imitation learning is to train a policy from an offline dataset through supervised learning. However, this approach suffers from a fundamental issue when the policy distribution cannot perfectly match that of the expert distribution, also known as the covariate shift problem. \cite{ross2010efficient}. Interactive imitation learning combats this issue through online expert intervention to recover from undesired states visited by the trained policy. 
One approach in interactive imitation learning is human-gated DAgger (HG-DAgger) \cite{kelly2019hg}, where a human supervisor decides when to correct a robot's actions. However, this requires continuous, one-on-one monitoring, making it resource-intensive and unscalable for parallel learning with multiple robots. The alternative method is called the robot-gated DAgger, which requires the robot to autonomously estimate its own uncertainty. A common challenge in interactive imitation learning is to maximize the utility of expert intervention while minimizing the burden on the expert. 

\subsection{Uncertainty Estimation of Robot Action}

The most widely used method for estimating action uncertainty is to use an ensemble of policies bootstrapped on a subset of the training data and using the level of disagreement among the policies output to estimate uncertainty \cite{menda2019ensembledagger, hoque2021thriftydagger, dass2022pato}. However, the ensemble method for uncertainty estimation can be problematic in practice, especially when the task allows multiple modes of completion. Specifically, they are prone to either mistaking in-distribution multi-modal data as out-of-distribution data or vice versa. This is because expressive ensemble policies may disagree due to the presence of multiple viable strategies where each policy is certain about its generated plan, leading to high action variance. Furthermore, if the policy generates an action sequence instead of a single action, the uncertainty estimation may explode due to divergence over time or lose accuracy if a shorter horizon is used for calculating action variance. 

Another uncertainty estimation method is to use a separately trained classifier to more explicitly detect OOD states using failure states \cite{liu2023model}. However, this requires significant efforts of collecting and labeling human-gated interventions as training data. In \cite{wong2022error}, the auxiliary reconstruction loss from the image observations is proposed to estimate uncertainty, but this method may be susceptible to confounding variables related to reconstruction complexity, potentially independent of the current state's in-distribution status.

In contrast to the existing robot-gated DAgger methods, our approach handles complex data distribution by incorporating the loss function of diffusion policy into the decision rule. This approach does not need multiple policies for uncertainty estimation, reducing both training and inference time while benefiting from using an expressive policy class.

\subsection{Policy Learning and Classification with Diffusion Models}
Diffusion models, originally proposed as generative models for images, have seen success in various decision-making tasks due to their ability to handle complex data distributions. In robotics, these models have been used in various aspects including reinforcement learning (RL) \cite{ada2024diffusion}, imitation learning \cite{chi2023diffusion}, task planning and motion planning \cite{carvalho2023motion}. Interestingly, diffusion models have also been shown to be effective zero-shot classifiers for the data distributions they are trained on \cite{li2023your, zhou2024adaptive, wang2024diffail}. This dual functionality of generative and discriminative capabilities enhances their value in decision-making tasks. Especially, it aligns well with a robot-gated DAgger framework, which requires both generation and evaluation of robot plans to decide when human intervention is necessary during task execution.

\section{Method}

\subsection{Problem Statement}

The goal of imitation learning is to collect a dataset and train a robot policy to complete a set of tasks with expert demonstrations. In the robot-gated DAgger scenario, the robot autonomously queries for expert intervention. We assume the expert is skilled in completing the task, which is considered successful if the goal is reached within the time limit. We assume access to success information. Our objective is to design an efficient robot query system that accurately predicts task failures, requesting expert intervention only when necessary.

\subsection{Diffusion Policy}
\subsubsection{Action Diffusion}
Diffusion Policy formulates action generation as a DDPM \cite{ho2020denoising}, where Langevin dynamics iteratively denoises a Gaussian prior, $\mathcal{N}(0,I)$, to recover structured actions. This approach enables stable training, handles multimodal action distributions, and has achieved state-of-the-art results in behavior cloning for robot manipulation \cite{chi2023diffusion}.

\subsubsection{Loss Function}

During training, Diffusion Policy samples a denoising timestep and a Gaussian noise vector to corrupt the action sequence. Given the corrupted actions and the original observation sequence, it predicts the target objective and optimizes using the following loss function:
\begin{align}
\label{diffusion1}
L_{\pi}(o,a,\epsilon,t) = 
\Vert f(\epsilon, a, t) - f_{\theta}(o, \sqrt{\alpha_t} a + \sqrt{1 - \alpha_t} \epsilon, t) \Vert^2 
\end{align}
where $o$ is the observation sequence, $a$ is the action sequence, and $\epsilon$ is a noise vector from Gaussian distribution $\mathcal{N}(0, I)$. The parameter $\alpha_t$ is a time-dependent factor that modulates the influence of the original data and the added noise during the diffusion process, and $t$ is the denoising timestep. The loss function $L(o, a,\epsilon,t)$ calculates the mean squared error between the prediction target  $f(\epsilon, a, t)$, which depends on the choice of prediction type, and the noise predicted by the model $f_{\theta}$, given the noised data at time $t$ \cite {salimans2022progressive}. 

\subsection{Diff-DAgger}
\label{ss:diff-dagger}

\subsubsection{OOD Detection via Diffusion Loss}


During robot rollout, the uncertainty of the current action plan is estimated using the diffusion policy's loss value for the given observation and the generated action. In order to establish a baseline, we calculate the expected diffusion loss $\mathscr{L}$ for each data point in the training dataset using equation \ref{eq:diffusion2}. 


\begin{equation}
\label{eq:diffusion2}
    \mathscr{L}(o,a, \pi) = \mathbb{E}_{\epsilon \sim \mathcal{N}(0,I), t \sim \mu(1,T)} L_{\pi}(o,a,\epsilon,t)
\end{equation}

For practical implementation, we select a batch size $N_{b}$ to sample both the noise and diffusion timesteps and compute the average loss over the batch.
Given a quantile threshold $\alpha$, we assume OOD if the diffusion loss of the generated action using Eqn \ref{eq:diffusion2} exceeds the $\alpha$-quantile of the expected diffusion losses from the training data. 
Additionally, expert supervision takes place only if $K$ past states consecutively violate the qauntile threshold $\alpha$. It helps distinguish brief fluctuations from persistent out-of-distribution cases, significantly reducing the rate of false positives.

\begin{algorithm}
\caption{Diff-DAgger}
\small 
\begin{algorithmic}[1]
\State \textbf{Require} Initial expert dataset $D_{exp}$, time-limit $T_L$, expert policy $\pi_{e}$, and thresholds $\alpha$ and $K$
\State \textbf{Initialize} diffusion policy $\pi_r$ by training on $D_{exp}$ 
\For{each trial}
    \State \textbf{Train} policy $\pi_r$ on dataset $D_{exp}$
    \State \textbf{Set} the threshold using $D_{exp}$ and $\alpha$
    \State control $\leftarrow$ robot
    \While {task is not done}
        \While {control = expert and task is not done}
            \State $a_e \gets \pi_e(o)$ \Comment{Expert provides action}
            \State Take action $a_{e}$
            \State $D_{exp} \gets D_{exp} \cup (o, a_{e})$ \Comment{Record state-action}
        \EndWhile
        \State $a_{r} \leftarrow \pi_{r}(o)$ \Comment{Policy provides action}
        \State loss $\leftarrow \mathscr{L}(o, a_{r}, \pi_{r})$ \Comment{Using Eq. \ref{eq:diffusion2}}
        \If {\textit{CDF}(loss) $> \alpha$ for last $K$ timesteps}
            \State control $\leftarrow$ expert \Comment{Expert takeover}
        \Else
            \State Take action $a_{r}$
        \EndIf
    \EndWhile
\EndFor
\end{algorithmic}
\label{algorithm:diff-dagger}
\end{algorithm}

\subsubsection{Data Aggregation and Policy Update}
We now summarize the data aggregation and policy update procedures. Diff-DAgger first initializes the policy $\pi_r$ by training it on N\textsubscript{i} initial expert demonstrations.
During rollout, the robot takes the actions in the environment until a high diffusion loss is detected. When the robot query is made, experts intervene from the current scene and complete the task. Upon completion of the task by either the robot or the expert, the task is initialized to a random initial configuration. If the robot neither queries nor completes the task after a time limit, automatic expert intervention takes place. After every N\textsubscript{d} expert intervention has been made, we perform a policy update, which consists of policy training and threshold setting. The policy training involves training the policy for a fixed number of epochs using the previously collected expert demonstration data, and threshold setting involves setting the base values by iterating through the training dataset to set the OOD violation threshold used for robot-query trigger.  
The full pseudo-code is available in Algorithm \ref{algorithm:diff-dagger}.

\section{Experiment}
We address the following three research questions to understand the effectiveness of Diff-DAgger.
\begin{compactenum}
    \item[\textbf{RQ1}] Can diffusion loss predict task failure (Section \ref{ss:RQ1})?
    \item[\textbf{RQ2}] Given a fixed number of expert interventions, can Diff-DAgger outperform existing methods (Section \ref{ss:RQ2})?
    \item[\textbf{RQ3}] Can Diff-DAgger reduce the wall-clock time of DAgger iterations (Section \ref{ss:RQ4})?
\end{compactenum}

\subsection{Experiment Setup}

\begin{figure}[t]
\centering
\includegraphics[width=0.48\textwidth]{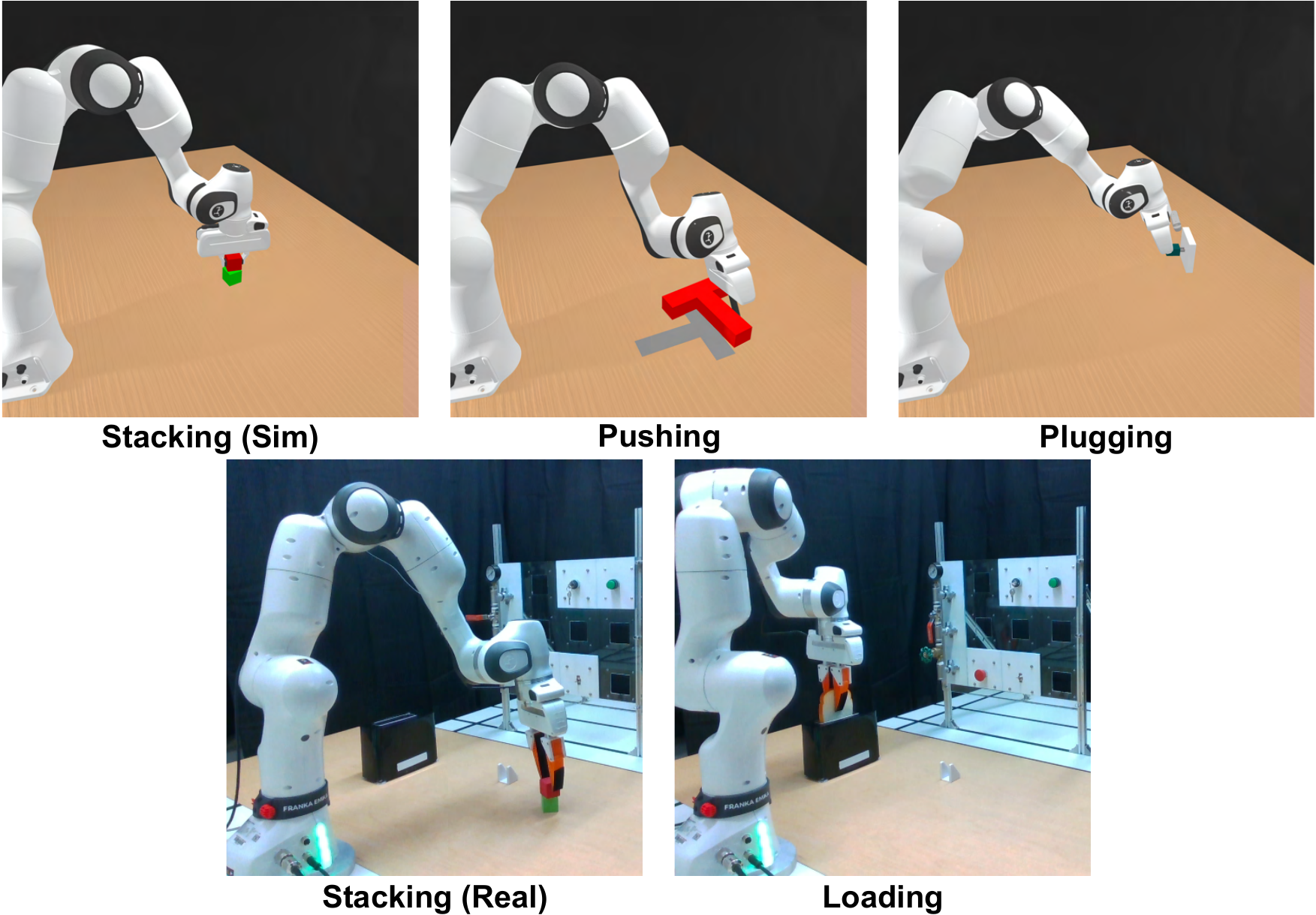}
\caption{Overview of simulation and real-world tasks: stacking, pushing, plugging, and loading.}
\label{fig:task_cam}
\end{figure}

\begin{table}[t]

\resizebox{0.485\textwidth}{!}{
\small
\setlength{\tabcolsep}{3.5pt} 
\renewcommand{\arraystretch}{1.3} 
\centering
\begin{tabular}{c | c | c c c c c c c c}
\toprule
Type & Task &  \makecell{\#\\Obj} & N\textsubscript{C} & \makecell{Contact\\Rich} & \makecell{High\\Prec} & T\textsubscript{L} & Expert Type & \makecell{Task\\Dur} & \makecell{Succ\\Rate} \\
\midrule
\multirow{3}{*}{Sim} & Stacking & 2 & 3 & False & False & 250 & Motion Planner & 60.3 & 1.00 \\
& Pushing & 1 & 1 & True & False & 250 & RL Agent & 65.7 & 0.93 \\
& Plugging & 2 & 3 & False & True & 500 & Motion Planner & 147 & 0.98 \\
\midrule
\multirow{2}{*}{Real} & Stacking & 2 & 3 & False & False & 600 & Human & 174.2 & 1.00 \\
& Loading & 2 & 3 & True & True & 600 & Human & 210.5 & 1.00 \\
\bottomrule
\end{tabular}
}

\caption{Task and Expert Summary. \# Obj: number of objects, N\textsubscript{C}: number of cameras, Contact Rich: contact-richness, High Prec: high-precision, T\textsubscript{L}: time limit in timesteps at 20 Hz in simulation or 30 Hz in real world, Task Dur: average expert task duration in simulation steps, Exp SR: expert success rate during intervention.}
\label{table:task_expert}
\vspace{-0.5cm}
\end{table}

\subsubsection{Tasks and Experts}

We test Diff-DAgger in the simulated and real robot environments.
We include three tasks in simulation—stacking, pushing, and plugging—using the ManiSkill environment \cite{taomaniskill3} and two tasks—stacking and loading—in real world. \cameraReady{In the stacking task, the robot picks up a red cube and places it on a green cube, evaluating its basic perception and manipulation skills. The pushing task requires using an end-effector stick to move a red tee to a predefined position, requiring multimodal planning and control of contact dynamics. The plugging task involves inserting a charger into a receptacle, testing fine manipulation and alignment accuracy. Lastly, the loading task requires the robot to pick up the soft bread and precisely load in onto a toaster, assessing the robot's ability to handle deformable objects with precision.}

For simulation tasks, we design experts as either multi-modal or long-horizon planners to evaluate different expert types. Specifically, we use RL agents and motion planners as the experts: the motion planner, is a long-horizon planner that generates actions based on future goals, whereas multiple RL agents with different reward functions provide different modes for task completion. For stacking and plugging tasks, we use a rule-based RRT motion planner to simulate the expert \cite{kuffner2000rrt}.
\cameraReady{For pushing, two experts are trained using PPO \cite{schulman2017proximal} with varying rewards: in addition to task-related reward, one agent is additionally rewarded for rotating the object clockwise direction, whereas other agent is rewarded for rotating counter-clockwise. During interactive learning, the agents alternate in providing demonstrations. We additionally compare learning from the counter-clockwise agent only. For real-world tasks, a proficient human expert teleoperated the robot using a VR controller.} The tasks and experts are summarized in Table \ref{table:task_expert}. 

\cameraReady{For the task setup, we use three cameras—left view, right view, and in-hand view—for pick-and-place tasks to mitigate visual occlusion, while a single top-down camera is used for the pushing task. In real-world experiments, we use a Franka Research 3 robot and use three RealSense D435 cameras, capturing 720p RGB images at 30 FPS, which is then downsampled to 256p at 15 FPS.}

\subsubsection{Baselines}

We compare Diff-DAgger to the following algorithms: Behavior Cloning, which uses the offline human demonstration data as the training data; Ensemble-DAgger, which uses five policies to calculate action variance for uncertainty estimation; and ThriftyDAgger \cite{hoque2021thriftydagger}, a SOTA robot-gated DAgger algorithm which incorporates risk measure into the decision rule using an offline RL algorithm. All methods use the diffusion policy with the same model architecture to ensure consistent expressiveness across methods.

 
\subsubsection{Model Learning}

\cameraReady{We adopt a model architecture similar to \cite{ronneberger2015u}, utilizing a U-Net \cite{ronneberger2015u} policy head with FiLM conditioning \cite{perez2018film}.}
We test both state-based and image-based data:
For the state-based case, we use the internal robot proprioception data (robot joint position and end-effector pose) and the ground-truth pose of the external objects as the input.
For the image-based case, we apply random crop augmentation following \cite{mandlekar2021matters} and extract the embedding vector using an R3M ResNet18 encoder \cite{nair2022r3m}. Instead of average pooling, we apply spatial softmax \cite{levine2016end} to the extracted features before concatenating them with proprioception data. Additionally, the observation horizon is set to one to reduce the amount of data processed in each batch. 
All approaches use position-based control, using either joint position control or end-effector pose control, \cameraReady{with either absolute or relative action representations \cite {chi2024universal}.} For end-effector pose control, a 6D representation for the orientation is used \cite{zhou2019continuity}.


\subsection{Predictability of Robot Query on Task Failure}
\label{ss:RQ1}
The effectiveness of a robot query system depends on its ability to respond to a failing episode. To quantitatively assess this, we evaluate it in simulation, where task outcomes can be efficiently recorded without the need for real-world rollouts. Specifically, we construct a confusion matrix between the robot’s query decisions and task outcomes:
For each task, we train the robot policy on a fixed number of expert demonstrations to roughly match 50\% success rate and evaluate for 100 randomly initialized episodes. We record task completion without expert intervention and whether the policy estimates high uncertainty during each episode, as described in Section \ref{ss:diff-dagger}. For each task, all methods use the same set of thresholds, including $\alpha$ and K, as shown in Table \ref{tab:hyperparam}. 

The results show that Diff-DAgger achieves a significant improvement over Ensemble-DAgger and Thrifty-DAgger in terms of its ability to predict task failure: Compared to Ensemble-DAgger, Diff-DAgger achieves 42.7\% higher F1 and 38.4\% greater accuracy scores on average across the simulated visuomotor tasks. Compared to Thrifty-DAgger, Diff-DAgger improves F1 scores by 22.6\% and accuracy by 32.8\%.
The results across tasks are shown in Table \ref{tab:prediction}.


\setlength{\tabcolsep}{4pt} 
\begin{table}[t!]
\centering
\resizebox{0.46\textwidth}{!}{
\begin{tabular}{>{\arraybackslash}M{1.2cm} >{\centering\arraybackslash}m{.6cm} >{\centering\arraybackslash}M{1.8cm} >{\centering\arraybackslash}M{0.6cm} >{\centering\arraybackslash}M{0.7cm} >{\centering\arraybackslash}M{0.55cm} >{\centering\arraybackslash}M{0.7cm}}
\toprule
Task         & N\textsubscript{D} & Method     & TPR & TNR & F1 $\uparrow$ & Acc $\uparrow$ \\ \midrule
\multirow{3}{*}{\makecell{\\Stacking\\}} & \multirow{3}{*}{\makecell{\\30\\}} 
             &           Ours       & 0.88   & 0.77     & \textbf{0.83} & \textbf{0.84}   \\ \cmidrule(lr){3-7}
             &      & Ensemble-DA   & 0.81   & 0.31   & 0.52 & 0.59   \\ \cmidrule(lr){3-7}
             &      & Thrifty-DA    & 0.95   & 0.21   & 0.54 & 0.65   \\ \midrule
\multirow{3}{*}{\makecell{\\Pushing\\(1 Expert)}} & \multirow{3}{*}{\makecell{\\40\\}} 
             &            Ours      & 0.83   & 0.92    & \textbf{0.88} & \textbf{0.85}   \\ \cmidrule(lr){3-7}
             &      & Ensemble-DA   & 1.00   & 0.11    & 0.60 & 0.73    \\ \cmidrule(lr){3-7}
             &      & Thrifty-DA   & 1.00   & 0.02    & 0.57 & 0.72   \\ \midrule
\multirow{3}{*}{\makecell{\\Pushing\\(2 Experts)}} & \multirow{3}{*}{\makecell{\\25/25\\}} 
             &            Ours    & 1.00   & 0.52    & \textbf{0.77} & \textbf{0.82}   \\ \cmidrule(lr){3-7}
             &      & Ensemble-DA   & 0.29   & 0.90   & 0.58 & 0.42    \\ \cmidrule(lr){3-7}
             &      & Thrifty-DA  & 0.58   & 0.66    & 0.62 & 0.62   \\ \midrule
\multirow{3}{*}{\makecell{\\Plugging\\}} &  \multirow{3}{*}{\makecell{\\100\\}}
             &            Ours    & 0.68   & 0.82    & \textbf{0.75} & \textbf{0.73}   \\ \cmidrule(lr){3-7}
             &      & Ensemble-DA   & 0.22   & 0.27   & 0.14  & 0.37   \\ \cmidrule(lr){3-7}
             &      & Thrifty-DA  & 0.26  & 0.31  & 0.22  & 0.21   \\ \bottomrule
\end{tabular}
}
\caption{Comparison of robot query decisions (query or no query) against task outcomes (success or failure) in 100 policy rollouts for visuomotor tasks.
Here, \textit{True Positive Rate (TPR)} refers to the proportion of queries made to failing episodes, while \textit{True Negative Rate (TNR)} refers to the proportion of non-queries made to successful episodes. N\textsubscript{D} is the number of expert demonstrations used for training the policy. 
}

\label{tab:prediction}

\vspace{0.3cm} 
\setlength{\tabcolsep}{1.7pt} 
\renewcommand{\arraystretch}{1.1} 
\centering
\resizebox{0.489\textwidth}{!}{
\begin{tabular}{lcccccccc|cc}
\toprule
& \multicolumn{2}{c}{\makecell{Stacking \\ (Sim)}} 
& \multicolumn{2}{c}{\makecell{Pushing \\ (1 Expert)}} 
& \multicolumn{2}{c}{\makecell{Pushing \\ (2 Experts)}} 
& \multicolumn{2}{c|}{Plugging}  
& \makecell{Stacking \\ (Real)} 
& {Loading} \\
\cmidrule(lr){2-3} \cmidrule(lr){4-5} \cmidrule(lr){6-7} \cmidrule(lr){8-9} \cmidrule(lr){10-10} \cmidrule(lr){11-11}
 & {State} & {Image} & {State} & {Image} & {State} & {Image} & {State} & {Image} & {Image} & {Image} \\
\midrule
BC           & {0.89} & {0.85} & {0.89} & {0.72} & {0.74} & {0.71} & {0.75} & {0.49} & {0.6} & {0.8} \\
Ensemble-DA  & {0.72} & {0.83} & {0.87} & {0.81} & {0.70} & {0.59} & {0.69} & {0.37} &  {-}  &  {-} \\
Thrifty-DA   & {0.79} & {0.86} & {0.85} & {0.81} & {0.69} & {0.68} & {0.65} & {0.47} &  {-}  &  {-} \\
Ours         & {\textbf{0.94}} & {\textbf{0.94}} & {\textbf{0.96}} & {\textbf{0.87}} & {\textbf{0.87}} & {\textbf{0.79}} & {\textbf{0.92}} & {\textbf{0.84}} & {\textbf{0.8}} & {\textbf{0.95}} \\
\bottomrule
\end{tabular}
}
\caption{Performance comparison across different tasks with state-based and image-based observations in simulation or real-world. We report success rates after the final DAgger iteration, using 100 evaluation episodes for simulation tasks and 20 for real-world tasks. For each task, the number of expert demonstrations is fixed (detailed in Table \ref{tab:hyperparam}) across methods. Our results show that Diff-DAgger consistently outperforms the other methods given the same number of demonstrations from the expert.}
\label{tab:results}

\vspace{0.3cm} 
\setlength{\tabcolsep}{1.5pt} 
\renewcommand{\arraystretch}{1.3} 
\centering
\resizebox{0.48\textwidth}{!}{
\begin{tabular}{lccccccccccc||c}
\toprule
Task & Action & T\textsubscript{o} & T\textsubscript{p} & T\textsubscript{a} & PredTg & T\textsubscript{d} & N\textsubscript{i} & N\textsubscript{f} & N\textsubscript{d} & $\alpha$ & K  & N\textsubscript{b} \\
\midrule
Stacking (Sim)    & Abs Pose & 1 & 32 & 8 & V-obj & 16  & 20 & 60  & 4 & 0.99 & 2 & 512 \\
Pushing (1 Expert) & Rel Joint & 1 & 32 & 2 & V-obj & 16  & 20 & 100 & 4 & 0.99 & 1 & 512 \\
Pushing (2 Experts) & Rel Joint & 1 & 32 & 2 & V-obj & 16  & 20 & 100 & 4 & 0.9 & 2 & 512 \\
Plugging          & Abs Pose & 1 & 64 & 8 & V-obj & 64 & 20 & 100 & 8 & 0.99 & 2 & 512 \\
\midrule
Stacking (Real)  & Abs Pose & 1 & 32 & 8  & V-obj & 16 & 40 & 120  & 20 & 0.99 & 2 & 512 \\
Loading          & Abs Pose & 1 & 64 & 8  & V-obj & 64 & 40 & 120  & 20 & 0.99 & 2 & 512 \\
\bottomrule

\end{tabular}
}
\vspace{0.04cm} 

\caption{Hyperparameters for Diffusion Policy and DAgger. Action: action space, T\textsubscript{o}: observation horizon, T\textsubscript{p}: prediction horizon, T\textsubscript{a}: rollout horizon, PredTg: prediction target when training diffusion policy, T\textsubscript{d}: number of diffusion timesteps for training and inference, N\textsubscript{i}: number of initial expert demonstrations, N\textsubscript{f}: final number of expert demonstrations, N\textsubscript{d}: number of expert intervention after each training session, K: patience for violation,  N\textsubscript{b}: Diff-DAgger specific batch size for calculating the diffusion loss.}
\vspace{-0.5cm}
\label{tab:hyperparam}
\end{table}

\subsection{Task Performance}
\label{ss:RQ2}
We evaluate the policy performance of methods for state-based and image-based observations. 
Diff-DAgger method consistently outperforms the baselines in task completion rate across tasks, assuming the same number of expert demonstrations is given. On average, Diff-DAgger achieves a 13.7\% improvement over the second-best method in state-based tasks.
In visuomotor tasks, including the real-world tasks, Diff-DAgger on average achieves a 25.2\% improvement over the other methods. 



\subsection{Wall-Clock Time for DAgger}
\label{ss:RQ4}
Diff-DAgger also significantly reduces wall-clock time for training, threshold setting and inference, in comparison to the Ensemble-DAgger-based methods. Whereas Ensemble-DAgger-based methods perform the ensemble action inference on the entire training data to set the numerical action variance threshold, Diff-DAgger requires passing a single batch per train data point to efficiently get the expected diffusion losses. Additionally, Diff-DAgger uses a single policy, reducing the wall clock time for both training and inference by the factor of the number of policies used for ensemble. Thrifty-DAgger extends Ensemble-DAgger by estimating risk, requiring extra data collection using the robot policy and additional risk estimator training, resulting in longer wall-clock time across tasks.

We demonstrate the results on the visuomotor pushing task, where Diff-DAgger reduced training time by a factor of 5.1, threshold setting time by a factor of 48, and inference time by a factor of 3.8 compared to Ensemble-DAgger, as shown in Fig \ref{fig:toy_example}. The overall reduction in average policy update time is by a factor of 7.8, reducing the total time to just 0.54 hours. Thrifty-DAgger requires additional data collection and training for risk estimation, and Diff-DAgger achieves corresponding reduction factors for training, threshold setting, and inference times of 6.4, 53, and 3.9, respectively.

\begin{figure}[t!]
\centering
    \includegraphics[width=0.98\linewidth]{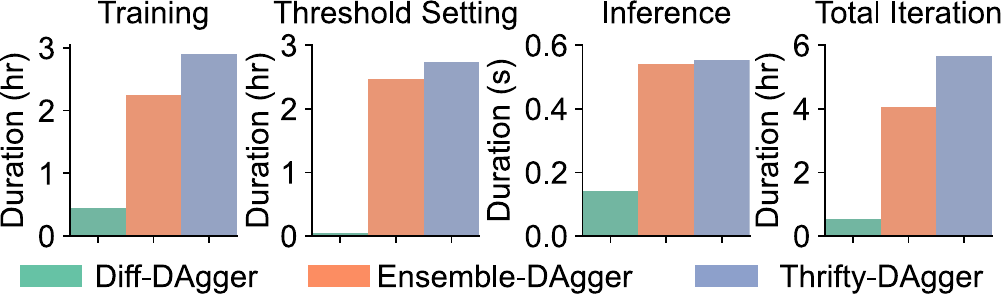}
    \caption{Average duration for training, threshold setting, inference, and total iteration for different methods, when training the final DAgger iteration of visuomotor policy for pushing task. A single A100 40 GB GPU is used for both training and inference.
    }
\label{fig:wallclock}
\vspace{-0.5cm}
\end{figure}


\subsection{Discussions}

\begin{figure}[t]
\centering
    \includegraphics[width=1.0\linewidth]{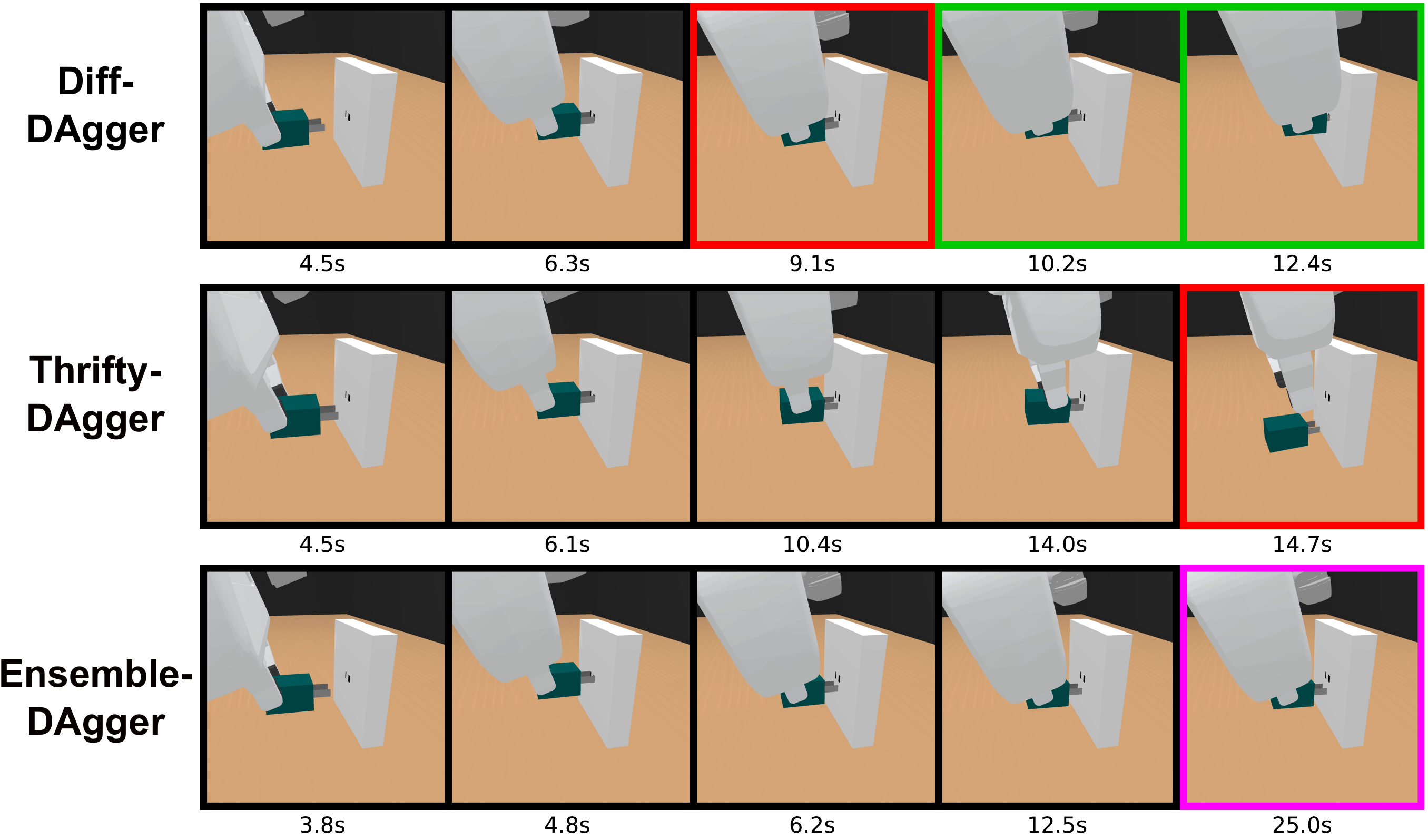}
    \caption{Robot query modes for different methods in the plugging task. Black boxes indicate policy rollout, red boxes represent robot queries, green boxes are expert interventions, and the purple box marks an automatic expert intervention due to timeout. In Diff-DAgger (\textbf{Top}), the robot promptly queries the expert when the charger slips at 9.1s, leading to a successful episode with expert help. In Thrifty-DAgger (\textbf{Middle}), the robot continues to try plugging in despite slipping, eventually querying for intervention after dropping it at 14.7s. In Ensemble-DAgger (\textbf{Bottom}), the robot fails to query despite not progressing, and times out at 25.0s to trigger automatic expert intervention. Expert interventions for Thrifty-DAgger and Ensemble-DAgger are omitted due to space.}
    \label{fig:behaviors}
    \vspace{-0.2cm}
\end{figure}

False positives and false negatives reported in Table \ref{tab:prediction} have distinct implications: a false positive means the robot unnecessarily asks for expert demonstration, simply losing the benefits of interactive learning and resembling offline imitation learning. In contrast, a false negative, where the robot overestimates its ability and fails, is more detrimental, as it can cause the robot to deviate significantly from the relevant states. 
\cameraReady{Both Ensemble-DAgger and Thrifty-DAgger incur a high rate of false negative in most tasks, especially when learning from two pushing experts. This is likely because the action variance present in the training dataset is high due to diverging plans (clockwise paths and counter-clockwise paths) given the same state, resulting in the policies' insensitivity toward true OOD data. And we speculate that this is the reason they perform worse than behavior cloning in many tasks. On the other hand, 
Diff-DAgger yields far fewer false negatives than other methods across tasks, making it a more suitable choice for an interactive learning framework with a complex data distribution. 
The} sensitivity to task failure is visualized in Fig \ref{fig:behaviors}, where we show robot query behavior from each DAgger method. For instance, Diff-DAgger proactively requests expert intervention once the robots start to lose grip of the charger, while the other methods fail to detect this issue—resulting in either dropping the object or getting stuck in one motion.

\cameraReady{It has been observed that the prediction type used for the diffusion model during training can have significant impact on both the convergence rate and the policy performance \cite {ze20243d}.
We find that sample prediction and V-objective prediction converges faster than epsilon and report the results using V-objective across the tasks as it works well considering both convergence rate and policy performance. }

\begin{figure}[t]
\centering
    \includegraphics[width=1.0\linewidth]{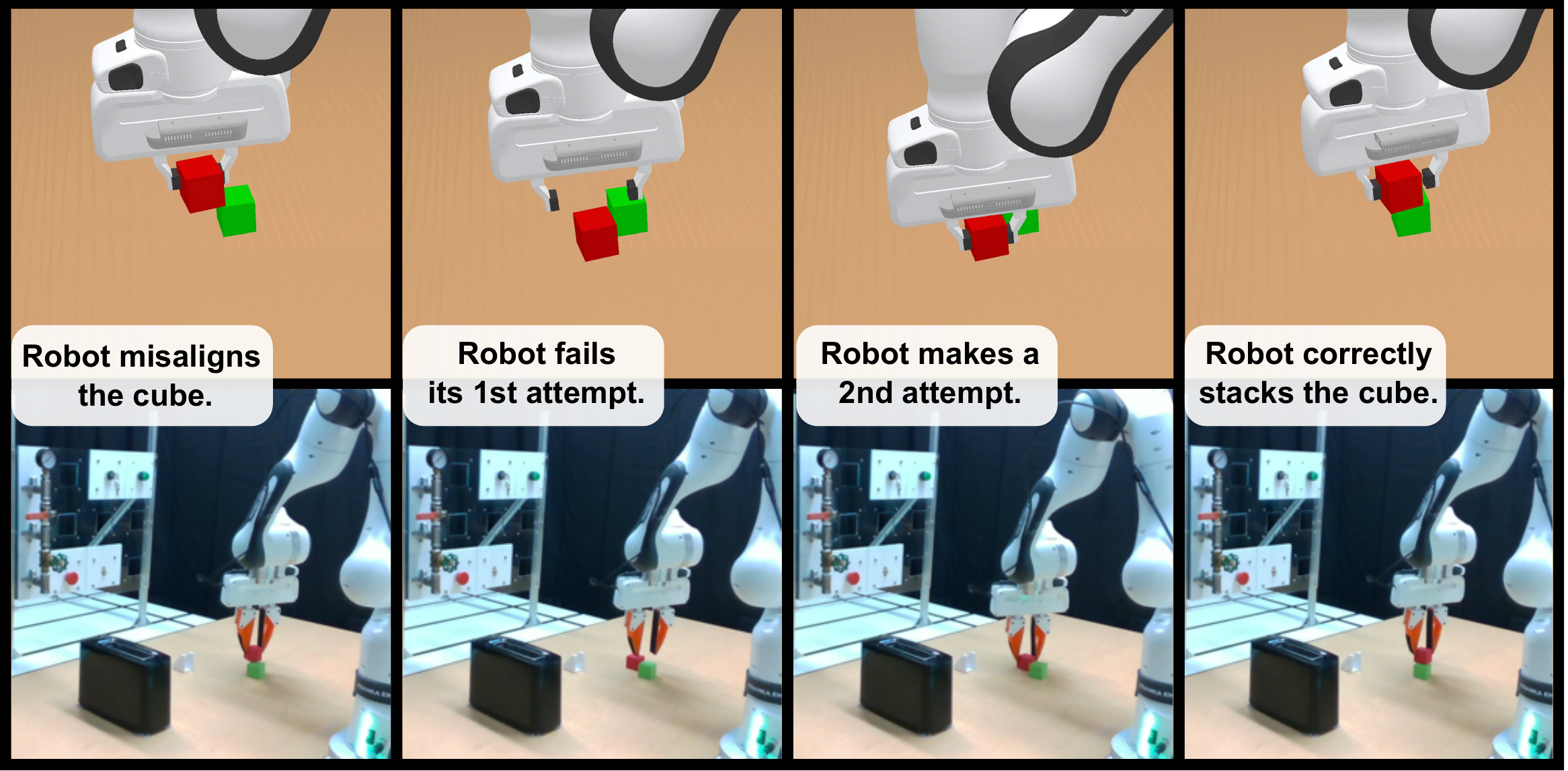}
    \caption{Emergent behavior of DAgger-trained policy for stacking. We observed an interesting new behavior with Diff-DAgger trained policies: the recovery behavior after initial failure. This explains the improvement in policy performance over policies trained using the offline approach.
}
\label{fig:emergent}
\vspace{-0.4cm}
\end{figure}

We also observe an interesting emergent behavior from policies trained on Diff-DAgger collected data: recovery behavior. As shown in Fig \ref{fig:emergent}, after the policy fails at its initial attempt to stack the cube, it subsequently tries again and completes the task. \cameraReady{This type of behavior is not typically observed unless carefully curated during data collection for offline behavior cloning.}

\section{Conclusion \& Future Work} 
\label{sec:conclusion}
In this work, we present Diff-DAgger, an efficient robot-gated DAgger method with diffusion policy.
While Diff-DAgger shows improvements in task failure prediction, policy performance, and wall-clock time for learning over other methods, there are limitations: 
Whereas we dramatically reduce the wall-clock time for DAgger iteration, using an expressive policy class such as diffusion policy comes at the cost of increased training time in general.
In addition, most of the choices we made to accelerate policy training is a trade-off between the policy performance, and we hope to address this in the future.
We hope that this work paves a way to open up more discussion for incorporating expressive policies into interactive learning frameworks in anticipation that computing will become faster over time. 

\section*{Acknowledgments}
This work was funded and supported by Delta Electronics Inc.
We acknowledge Research Computing at the University of Virginia for providing the computational resources and technical support.



\bibliographystyle{IEEEtranN}
\bibliography{references}

\clearpage

\end{document}